\documentclass{article}

\usepackage{PRIMEarxiv}

\usepackage[utf8]{inputenc}
\usepackage[T1]{fontenc}
\usepackage{url}
\usepackage{booktabs}
\usepackage{amsmath,amssymb,amsthm,amsfonts}
\usepackage{microtype}
\usepackage{fancyhdr}
\usepackage{graphicx}
\usepackage[round,authoryear]{natbib}
\usepackage[hidelinks]{hyperref}
\graphicspath{{./figures/}}

\DeclareMathOperator*{\argmin}{arg\,min}

\title{Skeletal Prototypes on Iterative Nerve Expansions}

\author{
  Jordan Eckert \\
  Department of Mathematics and Statistics \\
  Auburn University \\
  Alabama, USA \\
  \texttt{jpe0018@auburn.edu} \\
  \And
  Henry Schenck \\
  Department of Mathematics and Statistics \\
  Auburn University \\
  Alabama, USA \\
  \texttt{hks0015@auburn.edu} \\
}

\begin{document}
\maketitle

\begin{abstract}
    Prototype reduction replaces a training set with a smaller representation, and the established methods return a finite set of points. We propose Skeletal Prototypes on Iterative Nerve Expansions (SPINE). The model for each class is an embedded 1-complex rather than a point set. Its initial edge set is a class-conditional Mapper graph, so the data decide which localized clusters are joined. Later phases fit the vertices under a classification objective, and an observation is assigned to the class whose complex is nearest. The segments therefore enter the decision rule and not only the fitting. We evaluate SPINE on seventeen benchmark datasets under stratified 10-fold cross validation, against seven other prototype reduction methods at a matched budget. SPINE attains the highest mean accuracy and the best average rank. It is  significantly better than five of the seven competitors under Wilcoxon signed-rank tests with Holm correction. A budget sweep shows that the decision rule using the entire graph segments contribute most when prototypes are scarce, while the method as a whole competes best at moderate budgets. Construction cost places SPINE with the discriminative methods, and it is faster than generalized learning vector quantization on fourteen of the seventeen datasets.
\end{abstract}

\section{Introduction} \label{sec:intro} 

Prototype reduction methods aim to replace the training set with a much smaller representation (called a prototype set) that supports comparable classification accuracy at a fraction of the cost. The current literature is divided into methods that either select a subset of the observed data (\textit{instance} or \textit{prototype selection}) or construct new points (\textit{prototype generation}) \citep{bezdek_nearest_2001}. Prototype selection methods are organized by the type of selection performed \citep{garcia_prototype_2012}. Condensation methods retain points near class boundaries and discard interior points, edition methods remove noisy or mislabeled points, and hybrid methods do both. That is, condensation targets the reduction rate while edition targets the quality of the retained set, and the two goals are not generally served by the same rule. The same taxonomy separates methods further by the direction of search, such as incremental, decremental, batch, mixed, and fixed. Prototype generation methods are organized instead by the mechanism used to produce new points \citep{triguero_taxonomy_2012}. Positioning adjustment methods move a fixed number of prototypes under an update rule, as in the learning vector quantization (LVQ) family \citep{kohonen_self-organizing_1995}, while centroid-based and space-splitting methods merge points into averages taken within clusters or within regions produced by recursive partitions of the feature space.

Topological data analysis (TDA) summarizes a dataset by the shape of data, or its \emph{topology}. The best known tool is persistent homology, which records how features such as connected components, loops, and voids appear and disappear as a scale parameter grows \citep{carlsson_topology_2009}. A second construction, introduced by \citet{singh_topological_2007} called Mapper, takes a filter function defined on the data, covers the image of that filter by overlapping sets, clusters the observations falling in each cover set, and forms a graph whose nodes are those clusters and whose edges join clusters that share at least one observation. The resultant graph is the nerve of the pullback of the cover, and under mild conditions it approximates the Reeb graph of the filter \citep{carriere_statistical_2018}. The resolution of the summary is set by the number of cover sets and their percentage of overlap, meaning the level of detail is a choice rather than an outcome of the data. Mapper creates simple graphical representations of high dimensional data to uncover underlying structural patterns such as RNA hairpin folding intermediaries \citep{bowman_structural_2008}, identifying subgroups in biomedical data \citep{nicolau_topology_2011}, and analyze properties of neural networks such as their generalization capacity or expressivity \citep{ballester_topological_2024}. There has been a shift in recent work to treat the construction as something to fit in attempts to automate the selection of the lens \citep{oulhaj_differentiable_2024}, automatic covers such as in Ball Mapper \citep{dlotko_ball_2019} and G-Mapper \citep{alvarado_g-mapper_2025}, or using distribution guided probability models instead of fixed interval lengths \citep{tao_distribution-guided_2025}. 

The use of topology has entered the prototype reduction literature in two largely separate ways. The first uses proximity graphs as a device for deciding which points to keep. Editing rules built on Gabriel graphs and relative neighborhood graphs retain points whose graph neighbors carry a different label, since such points lie near a class boundary \citep{sanchez_prototype_1997}. Related constructions include local sets \citep{leyva_three_2015} and the class cover digraph family, in which prototypes are obtained as a dominating set of a proximity digraph built on the training data \citep{manukyan_classification_2020}. In each case the graph is built once, used to score or cover the points, and then discarded. The second way is through topology-preserving vector quantization. Self-organizing maps (SOM; \citealp{kohonen_self-organizing_1995}) and neural gas \citep{martinetz_topology_1994} place a fixed number of units by competitive learning while maintaining a neighborhood structure among them. Growing neural gas (GNG) extends this by inserting units over time and by adding edges through competitive Hebbian learning, so that an edge joins two units whenever they are the two units nearest to some observation \citep{fritzke_growing_1994}. The resulting edge set is a subgraph of the Delaunay triangulation restricted to the region the data occupy, and it therefore reflects the connectivity of the data and not only their density.

While the usage of topology has entered the supervised prototype reduction literature, we note the tools from TDA largely have not. Instead, TDA has entered the data reduction literature as a regularizer in dataset distillation \citep{li_fixed_2026}, as a sample-ranking score in data pruning \citep{roy_topoprune_2026}, and as a preservation objective in graph reduction \citep{chen_topograph_2026}. In the unsupervised setting, \citet{stolz_outlier-robust_2023} shows that the choice of landmark rule changes the recovered homology, as random selection favors dense regions while maxmin landmarks tend to include outliers and can produce loops that uniform random landmarks do not reproduce. In such cases, the retained set is judged by how well it reproduces a topological summary, not on classification accuracy as it is in prototype reduction \citep{garcia_prototype_2012, triguero_taxonomy_2012}. Thus TDA and prototype reduction have remained separate literatures. The separation is worth noting, since the proximity-graph editing rules and the induced connectivity of growing neural gas already indicate that the connectivity of a class carries information useful for reduction. 

\subsection{Previous Work and Contributions}
Every method described above returns a finite set of points, and the nearest-prototype rule then partitions the feature space into the Voronoi cells of that set. When points are used to represent a continuum, the number of points required to represent the space grows with the size of the region rather than with the difficulty of the problem. A model that carries edges as well as vertices does not share this limitation. An edge represents the segment joining its two endpoints, so a pair of stored points covers a continuum of locations at the cost of two. The nearest feature line rule uses this observation, classifying a query by its distance to the lines through pairs of same-class prototypes \citep{li_face_1999}. However, these lines are are a byproduct of the stored points, since every pair necessarily contributes one. The number of lines therefore grows quadratically in the number of prototypes a class holds, and nothing says which of them describes the class. What is needed is a small edge set that the data can choose. Mapper supplies such a structure directly. Applied to the observations of a single class, the construction returns a graph whose nodes are localized clusters and whose edges record which clusters share observations, so the graph describes how that class is connected at a chosen resolution. This differs from the connectivity produced by growing neural gas, which arises as a by-product of the quantization and is not used after the units are placed \citep{fritzke_growing_1994}. Fitting a  complex to data is itself an established idea, and the principal curve and principal graph literature has developed it over three decades \citep{hastie_principal_1989, krzyzak_piecewise_2002, albergante_robust_2020}. However those methods are unsupervised, and the complex describes a single point cloud under a representational objective.

We take the Mapper graph as the initial edge set of the class model. The construction decides which pairs of localized clusters are joined, at a resolution that grows with the size of the class rather than with a reduction budget. Everything after that is fitted to the data under a classification object. The resulting model for each class is an embedded 1-complex, that is, a skeletal network of vertices and the straight segments joining them, and an observation is assigned to the class whose complex is nearest. We call this construction Skeletal Prototypes on Iterative Nerve Expansions, or SPINE. Our contributions are as follows.

\begin{itemize}
  \item We propose SPINE, a prototype generation method in which the model for each class is an embedded 1-complex obtained from a class-conditional Mapper construction. To our knowledge, this is the first use of a Mapper construction for supervised prototype reduction.
  \item We evaluate SPINE on seventeen benchmark datasets under stratified 10-fold cross validation, against seven prototype reduction baselines drawn from both the selection and the generation families, at a matched prototype budget. 
  \item We sweep the prototype budget from $0.5\%$ to $25\%$. The sweep separates the contribution of using the skeletal prototype segments versus just the vertices. It also locates the operating range in which we recommend the method, which is at moderate budgets.
  \item We provide a public implementation and per-fold results for all methods and datasets considered in the corresponding author's GitHub.
\end{itemize}

The remainder of the paper is organized as follows. Section~\ref{sec:method} develops the SPINE construction phase by phase. Section~\ref{sec:results} reports the evaluation against seven prototype reduction methods at a matched budget. Section~\ref{sec:discussion} examines what the graph segments contribute across a budget sweep and what the construction costs. Section~\ref{sec:conclusion} concludes with limitations and directions for future work.

\section{Proposed Method} \label{sec:method}

\subsection{Mapper Background}
\begin{figure}[t]
  \centering
  \includegraphics[width=.75\linewidth]{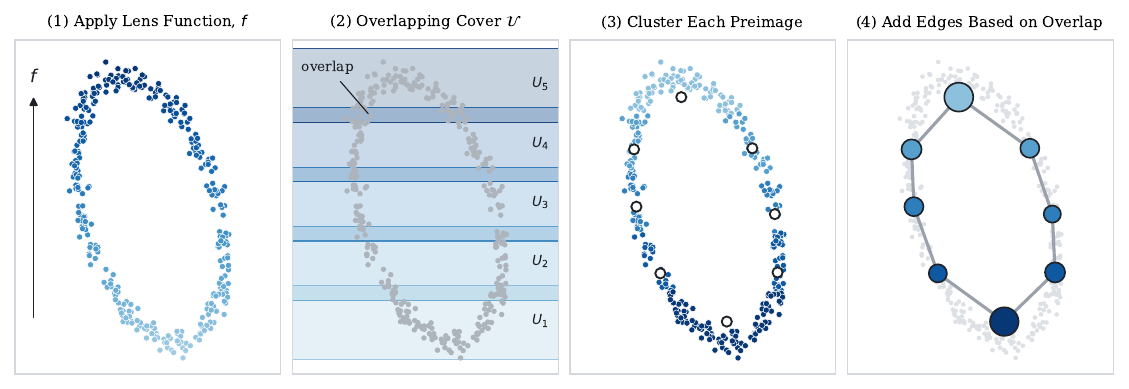}
  \caption{The Mapper construction on a noisy annulus of $n = 350$ points.
  (1) The lens $f$ is the leading principal direction; points are drawn in a rotated frame whose vertical axis is that direction and are
  shaded by their lens value. (2) The range of $f$ is covered by $5$ intervals of equal width, each
  overlapping its neighbors at a gain of $g = 0.25$; the overlaps appear as
  the darker bands. (3) The preimage of each interval is clustered in the
  ambient space, giving one cluster in $U_1$ and $U_5$ and two in each of
  $U_2$, $U_3$ and $U_4$. Open circles mark the cluster centroids. (4) Two
  clusters are joined when they share an observation. The resulting nerve
  has eight nodes, drawn at those centroids with area proportional to
  cluster size, and eight edges. We note that Mapper recovers the known topology of the
  annulus, $\beta_0 = 1$ and $\beta_1 = 1$.}
  \label{fig:mapper_example}
\end{figure}

Let $\mathcal{X} = \{x_1, \dots, x_n\} \subset \mathbb{R}^d$ denote a set of observations, and let $f : \mathcal{X} \to \mathbb{R}$ be a real-valued function called the \emph{lens}. Mapper summarizes $\mathcal{X}$ through the pair $(f, \mathcal{U})$, where $\mathcal{U} = \{U_1, \dots, U_\ell\}$ is a cover of the range of $f$ by $\ell$ overlapping intervals. The preimage $f^{-1}(U_k)$ collects the observations whose lens value falls in the $k$-th interval, and clusters that preimage in the ambient space. Two nodes in separate clusters are joined by an edge when they share at least one observation in the overlap. That is, the graph is the nerve of the cover of $\mathcal{X}$ obtained by pulling $\mathcal{U}$ back through $f$ and refining each preimage by the clustering. Two parameters set the resolution of the summary. The number of intervals $\ell$ controls how finely the lens is sliced. The gain $g$ sets the overlap, as each interval is widened to $(1+g)$ times the step, so consecutive intervals share a band equal to a fraction $g/(1+g)$. For a 1-dimensional lens and $g < 1$, only consecutive intervals meet, so the nerve contains no simplices of dimension greater than one and the summary is a graph. The Mapper graph is a discrete approximation of the Reeb graph of the lens. That is, it records how the level sets of $f$ split and merge, at the resolution the cover allows. For a Morse-type lens, the two agree up to features determined by where the cover's intervals fall relative to the critical values of $f$ \citep{carriere_structure_2018}. Figure~\ref{fig:mapper_example} illustrates the construction. 

Three properties of the Mapper graph carry into the method. First, the cover is what decides adjacency, so the initial graph joins only clusters that the cover placed side by side and carries no long-range connection. Second, the same graph is the lattice for the vertex placement of Section~\ref{sec:phase1}, where adjacency
in the graph means proximity in the data. Third, no later operation creates a cycle, so every cycle in the fitted model originally came from the Mapper graph. Each is established in the subsection that introduces it.

\subsection{The SPINE model}
\label{sec:spine-overview}

\begin{figure}[t]
  \centering
  \includegraphics[width=\linewidth]{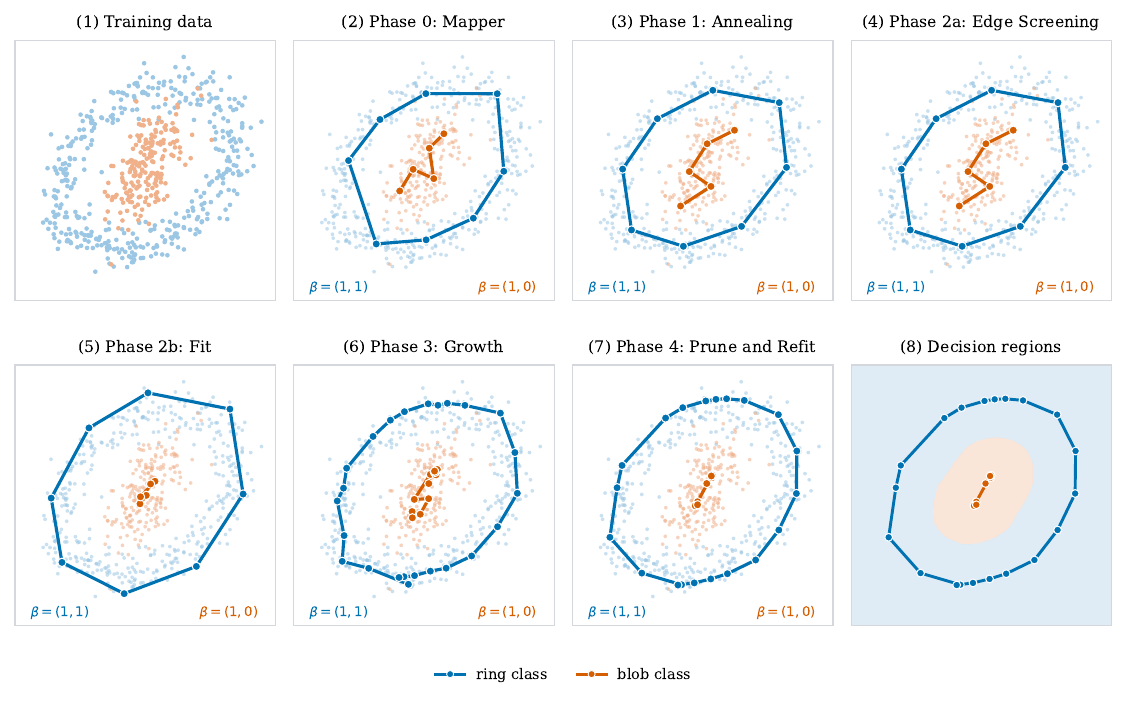}
  \caption{The phases of SPINE on a toy problem of 600 points, at a budget of 30 vertices apportioned as 20 to the ring class and 10 to the blob class. (1) The training data. (2) Phase 0 builds a Mapper nerve per class. (3) Phase 1 anneals the vertices onto the class while leaving the edges untouched. (4) Phase 2a screens the inherited edges. No edge deletion strictly decreased the validation risk on this toy problem, so the complex is unchanged here. (5) Phase 2b moves the vertices on the relative-distance margin, which contracts the blob's complex toward the region it must defend. (6) Phase 3 grows each class past its budget, alternating with refitting. (7) Phase 4 prunes to the exact budget and a final fitting pass follows. (8) The decision regions of the emitted model under $\hat{y}(x) = \argmin_c d(x, |S_c|)$. Betti numbers are printed in each panel in the class's own color, and the ring's $\beta_1 = 1$ is preserved throughout.}
  \label{fig:phases}
\end{figure}

SPINE replaces the finite prototype set with an embedded $1$-dimensional complex. Write $S_c = (V_c, E_c)$ for the complex of class $c$, where $V_c \subset \mathbb{R}^d$ is a set of vertices and $E_c$ a set of edges between them. We write $|S_c|$ for the realization of $S_c$, that is, the union of its vertices and the closed segments joining adjacent vertices. Betti numbers written $\beta_0(S_c)$ and $\beta_1(S_c)$ refer to the graph $(V_c, E_c)$, meaning its component count and its cycle count. For a graph these satisfy $\beta_1 = \#E_c - \#V_c + \beta_0$ where $\#$ refers to the cardinality. The complex is built in five phases, and each is described below. An observation is assigned to the class whose realization is nearest,
\begin{equation}
  \hat{y}(x) \;=\; \argmin_{c} \; d\bigl(x, |S_c|\bigr),
  \qquad
  d\bigl(x, |S_c|\bigr) = \min_{z \in |S_c|} \lVert x - z \rVert_2 .
  \label{eq:spine-rule}
\end{equation} The vertices are the prototypes, so reduction is measured in the same units as any other reduction method. Figure~\ref{fig:phases} gives a graphical representation of the phases applied to a toy dataset. 

\subsection{Phase 0: Initial Mapper Structure} \label{sec:phase0}

Before any fitting, SPINE reserves a stratified quarter of each training fold. Write $\mathcal{X}_{\mathrm{fit}}$ for the portion retained for fitting and $\mathcal{X}_{\mathrm{val}}$ for the reserved quarter, with $n_{\mathrm{fit}} = \#\mathcal{X}_{\mathrm{fit}}$. Write $\mathcal{X}_c \subset \mathcal{X}_{\mathrm{fit}}$ for the points of class $c$ in the fitting portion and $n_c = \#\mathcal{X}_c$, such that the class counts sum to $n_{\mathrm{fit}}$. Phase 0 initializes an independent Mapper graph on each class from that class's points alone, so supervision enters the construction only through the class split and not through the lens or the cover. The lens used is the projection onto the within-class first principal component (PC1). Projections of this kind are among the filter functions in common use \citep{madukpe_comprehensive_2025}. We specifically choose it on two grounds. The first is that PC1 introduces no constant of its own. Other common lens such as the Fiedler lens, feature projection, and eccentricity requires require a neighbor count, a declaration of which feature to use, or the full within-class distance matrix, respectively. The second ground is the geometry the method is built to exploit. PC1 is the direction in which the class spreads most. The class's points are $z$-scored before the lens is computed. That choice makes the lens invariant to any per-feature affine rescaling of the input. Features with zero within-class variance carry no within-class information and are left at zero. The sign of $f$ is fixed by a deterministic convention. The loading of largest magnitude is made positive, and ties are broken by index. Without such a convention the cover would be built on a range whose orientation is not reproducible. The $z$-scoring affects the lens alone. Every position computed downstream, including every vertex, lives in the coordinates of the training space.  

The cover of the range of $f$ is built from two constants, and they are set independently. The number of intervals follows the heuristic \begin{equation}
  \ell_c \;=\; \max\!\left\{2, \;\left\lceil (8 n_c)^{1/5} \right\rceil\right\},
  \label{eq:n-intervals}
\end{equation} which we adopt from the resolution rule derived by \citet{ahsan_maple_2026}. We use it as a heuristic and make no optimality claim for it here. What it supplies is a resolution growing like $n_c^{1/5}$, so that intervals do not starve as $n_c$ increases, while staying coarse enough that the nerve is not
over-refined. The gain is fixed at $g = 0.25$, and is never adapted per dataset or per interval. A varying gain would make adjacency mean different things in different places, while the nerve compares those adjacencies as commensurable, so the value is held fixed. Writing $f_{\min}$ and $f_{\max}$ for the extremes of $f$ on $\mathcal{X}_c$ and $\delta = (f_{\max} - f_{\min})/\ell_c$, the $k$th interval is
\begin{equation}
  U_k \;=\; \left[\,f_{\min} + \delta\!\left(k - \tfrac{1}{2}\right)
    - \tfrac{\delta(1 + g)}{2}, \;\;
    f_{\min} + \delta\!\left(k - \tfrac{1}{2}\right)
    + \tfrac{\delta(1 + g)}{2} \,\right],
  \qquad k = 1, \dots, \ell_c ,
  \label{eq:cover}
\end{equation} so consecutive intervals share a band of width $\delta g$ and non-consecutive intervals do not meet. The sampling results of \citet{carriere_statistical_2018} holds at a gain below one half, which bounds the gain from above. A failure also waits at the other end, since the overlap is what creates edges and a cover that barely overlaps returns a nerve with too few edges. We fixed $g = 0.25$ before any experiment, and elected to not tune it on the evaluation suite; we offer no principle that selects the value other than we believe it to be between both bounds. As such, its sensitivity is untested and we refer selection of an theoretically optimal gain for future work in Section~\ref{sec:conclusion}. Consecutive intervals here share a fifth of as opposed to the one-half in \citet{ahsan_maple_2026}, thus we take the interval count as a fixed heuristic, rather than as a quantity tied to the gain. 

Intervals are placed uniformly on the lens itself. An interval holding too few observations cannot resolve two clusters even in principle. When any interval of the uniform cover falls below that population, the cover is laid instead on the lens pushed through its own empirical distribution function. That reparameterization is monotone. Ties take the average rank, so points sharing a lens value are never placed in different intervals. It changes how the range is divided, but not the order of the points along the lens. Uniform placement is the default because the reparameterization buys resolution in dense regions by spending it in sparse ones. Sparse regions are disproportionately where branch points and connecting necks lie.

The points falling in each interval are clustered in the ambient space under the Euclidean metric with HDBSCAN \citep{hutchison_density-based_2013} at constants that are global and identical in every interval of every class. The minimum cluster size $\kappa$ and the minimum sample count $\lambda$ are \begin{equation}
  \kappa \;=\; \left\lfloor \min\left\{ 100, \; \max\left\{ 15, \;
    0.005\, n_{\mathrm{fit}} \right\} \right\} \right\rfloor,
  \qquad \lambda \;=\; 5 .
  \label{eq:mcs}
\end{equation} The default ties $\lambda$ to $\kappa$, which would otherwise vary depending on dataset. In this way $\kappa$ and $\lambda$ decide what counts as structure, while every point contributes to connectivity. Cluster selection uses excess of mass with single clusters permitted. Permitting single clusters is deliberate as one cluster is the correct answer for an interval meeting a single connected component. The rate within it, and the sample count, were fixed before an experiment and were not tuned on the evaluation suite. We make no claim for their optimality, and recommend practitioners to tune both under cross-validation if applying SPINE in practice. 

HDBSCAN labels low-density points as noise, and inside a Mapper cover such a point would join no cluster and contribute to no edge. Each vertex is placed at the centroid of its cluster's non-noise labeled members. Discarding noise would remove connectivity preferentially where the structure is decided. Noise points are therefore assigned to their nearest cluster within the same interval, for edge computation only. An interval that HDBSCAN labels as all noisy throughout is treated as a single cluster. 

\subsection{Phase 1: Annealed Representation} \label{sec:phase1}

The original placement of Phase 0 vertices are constrained by the Mapper algorithm's placement. The vertices are annealed in Phase 1 until they describe the class, with nothing else about the nerve changed. Phase 1 runs per class using competitive learning to update the vertex positions similar to SOM. The lattice is the nerve graph itself rather than a fixed grid. Write $w_1, \dots, w_m$ for the vertex positions of one class, and $d_G(i, j)$ for the hop distance between vertices $i$ and $j$ in the graph $(V_c, E_c)$. At epoch $t = 0, \dots, T-1$ the two schedule parameters are \begin{equation}
  \sigma_t \;=\; \sigma_0 \left(\frac{\sigma_T}{\sigma_0}\right)^{t/(T-1)},
  \qquad
  \varepsilon_t \;=\; \varepsilon_0
    \left(\frac{\varepsilon_T}{\varepsilon_0}\right)^{t/(T-1)} ,
  \label{eq:phase1-schedule}
\end{equation} so that both decay geometrically from their initial to their final values. The points of the class are then presented once each, in a random order, and
for each presented point $x$ the winning vertex is
$j^{\star} = \argmin_{j} \lVert x - w_j \rVert$ and every vertex is moved by
\begin{equation}
  w_j \;\leftarrow\; w_j \;+\;
    \varepsilon_t \, \exp\!\left(-\frac{d_G(j^{\star}, j)}{\sigma_t}\right)
    \left(x - w_j\right) .
  \label{eq:phase1-update}
\end{equation} The updates are online rather than batched. That is, each point is matched against the positions the previous point left behind. For disconnected pairs, no hop distance exists. Hop distances between vertices in different connected components are set to $m$, which exceeds any hop distance realizable in a simple graph on $m$ vertices.

Equation~\eqref{eq:phase1-schedule} and its constants follow neural gas. \citet{martinetz_topology_1994} anneal every schedule parameter geometrically from an initial value to a final one across the run, which is the form used here. They set the adaptation rate to run from $0.3$ to $0.05$. We use $0.4$ to $0.02$, which is the same form over a wider span. Their neighborhood range runs from $0.2N$ to $0.01$, where $N$ is the number of units. Two differences keep those two values from transferring directly. Neural gas weights a unit by its rank among all units by distance, while $\sigma_t$ is a hop distance on a fixed graph. Their initial range also scales with the model, and ours does not. What does carry over is the shape of the convention. The initial range is a small fraction of the model size, and the final range is below one. To find $\sigma_0$, we applied Phase 0 to each dataset in the evaluation suite. The median vertex count per class was six. Their rule at $m = 6$ therefore gives $0.2m = 1.2$. We set $\sigma_0 = 2$, which is wider than the convention gives. The coupling exists to move the nerve as a connected object. At $\sigma_0 = 2$ the weights are $0.61$, $0.37$, and $0.22$ at one, two, and three hops. The measurement itself uses Phase 0 alone, which sees no test partition and no accuracy,
so it is a structural calibration rather than a fit. We recommend practitioners repeating this calculation for $\sigma_0$ before fitting, though we elect not to do for each individual datasets when presenting results here. The final width has to make the coupling numerically absent by the last epoch, and $\sigma_T = 0.05$ gives a one-hop weight of $\exp(-20)$. Neural gas advances its schedule at every presentation, over $200N$ of them, whereas we advance once per epoch, so each parameter takes thirty distinct values.

We use $T = 30$ epochs as a heuristic, and the value was never swept. We note the exponent in Equation~\eqref{eq:phase1-schedule} is $t/(T-1)$, so the shape of the anneal does not depend on $T$. What $T$ sets is how many times each point is presented, and how finely the schedule is sampled. The positions are not run to a fixed point, since the rate ends at $\varepsilon_T = 0.02$, rather than at zero. The natural objection to quantizing along a graph is that a constrained quantizer cannot beat an unconstrained one. The objection does not bind here, since the constraint is transient. The coupling is intended to buy a better basin of attraction and to withdraw before the solution is settled. The constants of Equation~\eqref{eq:phase1-schedule} were fixed before the campaign, and their sensitivity is untested. A class whose nerve has a single vertex is a degenerate case, and the vertex is set to the class mean, which is the fixed point the expected update would reach. 

\subsection{Phase 2}
\subsubsection{Phase 2a: Edge Screening}
\label{sec:phase2a}

Under the previous phase, the vertices were moved with the edges intact, however the location of the edge to the new vertices might not provide a good fit to the data under the decision rule which measures distance to segments. The new vertex locations could possible draw the edges of $S_c$ through empty space, and attracts queries that belong to another class. Phase 2a screens the inherited edges and deletes the ones that do not earn their place. Vertices are never added or deleted during this phase, and no edge is every created. 

The default criterion is validation risk, and two criteria based on segment support are also implemented, together with a composite that runs the support test and then the risk test. Let $\mathcal{X}_{\mathrm{val}}$ be the held-out portion of the training fold, and write \begin{equation}
  R \;=\; \frac{1}{\#\mathcal{X}_{\mathrm{val}}}
    \sum_{(x,y) \in \mathcal{X}_{\mathrm{val}}}
    \mathbf{1}\!\left\{\hat{y}(x) \neq y\right\}
  \label{eq:val-risk}
\end{equation} for the risk of the current model under the decision rule of Equation~\eqref{eq:spine-rule}. Within each class the edges are visited in order of decreasing length, and each is deleted only when its deletion \emph{strictly} decreases $R$. Class are visited in index order. The pass is greedy, in that an accepted deletion becomes the baseline against which the remaining edges are judged. The strictness is load-bearing rather than stylistic. On a finite validation set, $R$ takes values in multiples of $1/\#\mathcal{X}_{\mathrm{val}}$, so exact ties between the current model and a trial deletion are common. A permissive rule, one that deletes on improvement or equality, therefore discards edges by numerical accident. The segments are the hypothesis class, therefore a rule that deletes on ties can potentially strip a nerve graph back to isolated points and collapse the method into a point-prototype classifier. 

Because this phase only deletes, the graph can only lose connection relative to the nerve. A component the nerve found connected may split, and a cycle it found closed may break. No later operation can join two components or close a cycle. The model thus never carries a cycle the nerve did not have. The component count is bounded in the same direction, with one exception. When the budget leaves Phase 4 no removal that preserves both Betti numbers, it removes a vertex regardless, and removing an isolated vertex deletes a component outright. We record the Betti numbers of every class here and again at emission, together with the numbers of forced removals, so that a run reports what its later phases did rather than assuming it.  

\subsubsection{Phase 2b: Discriminative Fitting}
\label{sec:phase2b}

Phase 2b moves the vertices with reference to the other class to reduce classification error. Only the vertex positions change, so the vertex count and the edge set are constant throughout this phase. The criterion here is the relative-distance margin of the generalized learning vector quantization (GLVQ; \citealp{sato_generalized_1995}) family. The gradient fitting uses the portion of the training fold served for fitting, rather than the held-out split. The distance to a prototype is replaced by the distance to a realized complex. For a training point $x$ with label $y$, write \begin{equation}
  d^{+} = d\bigl(x, |S_y|\bigr), \qquad
  d^{-} = \min_{c \neq y} d\bigl(x, |S_c|\bigr), \qquad
  \mu = \frac{d^{+} - d^{-}}{d^{+} + d^{-}} .
  \label{eq:margin}
\end{equation} Then $\mu \in [-1, 1]$, and $\mu < 0$ exactly when $x$ is strictly closer to its own class than to any other. The objective is the smoothed error
\begin{equation}
  L \;=\; \sum_{(x,y)} \phi(\mu), \qquad
  \phi(\mu) = \bigl(1 + \exp({-\theta \mu})\bigr)^{-1} ,
  \label{eq:loss}
\end{equation} Because $\mu$ is bounded in $[-1, 1]$, a scale near one leaves the per-point weight $\phi'(\mu)$ nearly constant, which is the linear variant \citet{sato_generalized_1995} report as the weaker of the two. We fixed $\theta = 3$ as a heuristic that avoids that flat regime without concentrating the weight on a narrow band at the boundary and do not tune it during evaluations. 

The gradient follows from the envelope theorem. Let $p$ be the point of $|S_c|$ nearest to $x$. The projection is held fixed while the vertices move. When $p$ lies on the segment between vertices $u$ and $v$ at projection parameter $\tau \in [0,1]$, so that so that $p = u + \tau(v - u)$, \begin{equation}
  \frac{\partial d}{\partial u}
    = -(1 - \tau)\,\frac{x - p}{\lVert x - p \rVert},
  \qquad
  \frac{\partial d}{\partial v}
    = -\tau\,\frac{x - p}{\lVert x - p \rVert},
  \label{eq:seg-grad}
\end{equation} and, writing $s = d^{+} + d^{-}$, $\partial \mu / \partial d^{+} = 2 d^{-} / s^{2}$ and $\partial \mu / \partial d^{-} = -2 d^{+} / s^{2}$. Composing the three factors of $\partial L / \partial u$, a vertex $u$ of the same-class complex moves by \begin{equation}
  \Delta u \;=\; \eta \, \phi'(\mu) \,
    \frac{2 d^{-}}{s^{2}} \, (1 - \tau) \,
    \frac{x - p}{\lVert x - p \rVert} ,
  \label{eq:raw-step}
\end{equation} with $\eta = 0.05$ selected as the convention. A vertex of the nearest wrong-class complex moves by the same expression with $2 d^{+}/s^{2}$ in place of $2 d^{-}/s^{2}$ and the opposite sign. A query therefore pulls on its nearest segment and splits the correction between the two vertices anchoring it, weighted by where it projects. A segment counts as nearest only when it beats the nearest vertex outright. A projection falling at an endpoint ties that vertex rather than beating it, so the update is taken at the vertex alone and reduces to the GLVQ update on points. 

Only the vertices incident to an edge or vertex that is the nearest wrong-class object for at least one training point are moved. The decision boundary is the set of points equidistant from two complexes of different classes, so a vertex that is never the nearest wrong-class object therefore cannot move it. The boundary set is obtained with no additional cost, since the $d^{-}$ winners observed while sweeping one epoch define the set used by the next. A single forward pass bootstraps the first epoch. The restriction applies to vertices and not to edge segments, so an edge with one interior endpoint moves only its boundary end. 

The step length needs separate treatment. The gradient in Equation~\ref{eq:seg-grad} carries a factor of $1/\lVert x - p \rVert$, which is an artifact of taking the margin through unsquared distances. As a result, the step diverges as the model comes to fit the data well. It does not arise in the form \cite{sato_generalized_1995} derive which uses squared distances. We recover the correct behavior by preconditioning each side by the distance of the object being moved. Since $\partial (d^{2}/2) / \partial v = d \cdot \partial d / \partial v$, the preconditioned step is exactly the gradient taken through the squared distance. Each vertex thus moves along its own gradient direction at a positive multiple of its length, so descent is preserved. The surviving coefficient is $2 d^{+} d^{-} / s^{2}$, which is symmetric in the two
cases and bounded by $1/2$, and $\phi'$ is bounded by $\theta/4$, so both steps obey
\begin{equation}
  \lVert \Delta u \rVert \;\leq\; \frac{\eta \, \phi'(\mu)}{2}
  \;\leq\; \frac{\eta \theta}{8} .
  \label{eq:step-bound}
\end{equation} Because the step no longer shrinks as the fit improves, the learning rate alone sets how finely the model can settle. 

Phase 2b runs after screening, after each round of grown in Section~\ref{sec:phase3}, and after the pruning of Section~\ref{sec:phase4} so that the schedule alternates between fitting and growth. That is, the geometry is refit after every step that changes the vertex positions since leaving them would place them in positions fitted to a different model. Each run is given a fixed number of epochs rather than a convergence test, which were selected a priori, and their sensitivity is untested.

\subsection{Phase 3: Attribution-Driven Growth} \label{sec:phase3}

Nothing so far has been sized to the a priori selected prototype budget. Write $M_c$ for class $c$'s share of the budget $M$. The vertex count $m$ arriving in Phase 3 is set by
the cover of Section~\ref{sec:phase0}, and it need not equal $M_c$. The remaining two phases are used in conjunction to obtain the reduction budget. The remaining two phases are used in conjunction to obtain the reduction budget. Phase 3 runs per class, and it adds vertices when necessary to where the current model fails. Exactly two operations, those of the elastic principal graph in \citet{albergante_robust_2020}, are permitted which grows a complex by bisecting an edge and by attaching a node to node. We adopt them specifically because they are the two moves that add a vertex to a $1$-dimensional complex without changing its homotopy type. A site is chosen in three steps. The model is evaluated on the held-out split and the misclassified points of class $c$ are collected, each error is attributed to the object of $S_c$ nearest to it, and the site is the object carrying the most attributed errors. Ties are broken by a fixed ordering so that the choice is deterministic, and the errors are recomputed after every addition. The operations are: \begin{itemize}
  \item \emph{Bisect.} An edge $(u, v)$ is replaced by $u - w - v$, with the
    new vertex $w$ at the centroid of the errors attributed to that segment.
    The centroid generally does not lie on the segment, and this is how the
    complex acquires curvature.
  \item \emph{Extend.} A new vertex $w$ is attached to an existing vertex
    $u$ by a single edge, at the centroid of the errors attributed to the
    winning site.
\end{itemize}

The operation selection is decided by where the errors sit. A site that is itself a vertex extends from that vertex, since the errors are already piled at a point of the complex. A site that with equal errors along an edge segment bisects. If the majority of errors along an edge segment project to one endpoint, and that endpoint is a leaf, the operation extends from the leaf. That is, errors in a segment's interior ask for curvature, while errors piling at a free end ask for reach.

Each class is grown past its budget on purpose. Phase 4 then removes the surplus by validation risk, so the overshoot is what gives the pruning pass a choice to make. The growth is toward a cap of $\lceil 1.2 \, M_c \rceil$ vertices, where $M_c$ is its share of the budget, capped at the number of points the class has in the fitting portion of the fold. The alteration between Phase 3 and Phase 4 runs for three rounds, and in each round a class receives the larger of three vertices and an equal share of its remaining shortfall, after which Section~\ref{sec:phase2b} refits the model. Holding the number of alternations fixed and letting the chunk size scale keeps the rhythm of fitting and growth the same at every budget.

Held-out errors are a finite supply, and a class can exhaust them before it has filled its growth budget. This happens when the class is easy to separate, and it happens by arithmetic when the budget approaches the size of the held-out split. Attributed errors then cannot furnish a site for every vertex, even if the model misclassified every held-out point. The site is then chosen by representation error instead, as the object carrying the greatest sum of squared distances from the same-class training points assigned to it, with the new vertex at their centroid. The same two operations apply, so only the error being reduced changes. This is the one place in the phase where training points rather than held-out points drive a decision, and every use is counted in the provenance. Growth therefore turns to representation once the held-out errors are spent. 

\subsection{Phase 4: Pruning to the Budget} \label{sec:phase4}

Phase 4 removes vertices until each class holds exactly its share of the budget. Vertices go one at a time, and at each step the one removed is whichever leaves the validation risk of Equation~\eqref{eq:val-risk} lowest, with ties going to the first such vertex encountered. Classes are pruned independently, with the other class(es) being held fixed. Deleting a vertex $v$ together with its incident edges gives
\begin{equation}
  \Delta \beta_1 \;=\; \Delta \beta_0 - \bigl(\deg(v) - 1\bigr) ,
  \label{eq:removal}
\end{equation} where $\deg(v)$ is the number of edges incident to $v$, so both Betti numbers survive a plain deletion only at $\deg(v) = 1$. The permitted removals are therefore the inverses of the two operations of Section~\ref{sec:phase3}. \begin{itemize}
  \item A vertex of degree one may be removed, which undoes an extension. The vertex and its single edge leave, and its neighbor carries the component.
  \item A vertex of degree two whose neighbors are distinct  and not already adjacent may be removed, which undoes a bisection. Two edges leave and one arrives, so the deletion is not plain and both numbers are preserved.
\end{itemize} Everything else is excluded. A vertex of degree zero is a component of its own, so removing it lowers $\beta_0$. A vertex of degree two whose neighbors are already adjacent cannot take the replacement edge, since the complex is a simple graph, and the removal reverts to Equation~\eqref{eq:removal} with $\beta_1$ falling by one. A vertex of degree three or more cannot be removed at all, since Equation~\eqref{eq:removal} admits no $\deg(v) \geq 3$ at which both differences
vanish. 

Some configurations admit no permitted removal. A forced removal is taken by the same least-risk criterion to match the budget. The removal is either from the vertices of degree at most two, or from all vertices when that set is empty as well. Equation~\eqref{eq:removal} bounds the damage. Deleting a vertex $v$ splits its component into at most $\deg(v)$ pieces, so $\Delta \beta_0 \leq \deg(v) - 1$ and $\beta_1$ can never rise. A forced removal can lose a cycle, then, but it can never invent one. It can also lower $\beta_0$, but only by removing an isolated vertex and with it a whole component. Every forced removal is counted and reported. We note that forced removals occur only in severely imbalanced minority classes or because the cover of Section~\ref{sec:phase0} (which is fixed before the budget is known) is already above the budget target. 

\section{Performance Evaluation} \label{sec:results}

In this section, we evaluate SPINE against seven prototype reduction methods, namely stratified random sampling (Random) and selection of prototypes by greedy optimal transport (SPOT; \citealp{gurumoorthy_spot_2021}),  reduction by space partitioning (RSP3; \citealp{sanchez_high_2004}), per-class $K$-means (KMeans), LVQ3, GLVQ, and GNG. The first two are prototype selection methods, and the remaining are prototype generators. 

RSP3, LVQ3 and SPOT are Python reimplementations from their source papers, GLVQ uses the \texttt{sklvq} \citep{veen_sklvq_2021} library at its own defaults, and GNG is vendored from Guille's public implementation \citep{guille_growingneuralgas_2016} and repaired to run on the current stack. GNG grows a network until it reaches the allotted size. Edge ageing can prune units below that size, so we insert further units by the algorithm's own rule until the count is budget exact. The number of passes is derived from the insertion schedule rather than tuned, since the schedule must be able to reach the quota. Each fold's seed is derived from a fixed base seed so that the stochastic methods are reproducible run to run. 

\subsection{Experimental Setup}

We use seventeen numeric classification datasets ranging from 178 to 20{,}000 observations, 2 to 60 features, and 2 to 26 classes. For the fifteen KEEL datasets we use KEEL's own published ten-fold cross-validation partitions. EEG Eye State and Waveform have no published partitions, and we use a seeded stratified 10-fold split at the same fold count. Features are standardized with the scaler fit on each training fold alone and applied unchanged to the corresponding test fold. Min-max normalization fixes each feature's range, and the range is fixed by the extremes. A single outlier therefore compresses the bulk of the data into a small part of the unit cube, which is where a margin gradient behaves the worst. 

RSP3 is parameter-free, deterministic, and blind to accuracy \citep{sanchez_high_2004}. Its prototype count therefore depends on the training fold alone. RSP3 runs first in every fold, and its count becomes the budget for SPINE, KMeans, Random, LVQ3, SPOT, GLVQ, and GNG. Table~\ref{tab:datasets} details the datasets and matched budget constraint. Each method needing a per-class budget apportions the matched total over the training-fold class frequencies by largest remainder, with a floor of one prototype per class. SPOT is the exception, since it selects globally and carries no per-class quota \citep{gurumoorthy_spot_2021}. 

\begin{table*}[t]
\centering
\small
\caption{The seventeen datasets, ordered by size. $M$ is the number of prototypes RSP3 produces on the training fold for the matched budget. Every budget-matched method emits exactly $M$ prototypes, so the reduction rate applies to all of them. Prototype counts are means over the ten outer folds, rounded to the nearest integer.}
\label{tab:datasets}
\begin{tabular}{lrrrlrr}
\toprule
Dataset & $n$ & $d$ & Classes & Source & $M$ & Red. \\
\midrule
Wine          &   178 & 13 &  3 & KEEL   &   40 & 0.753 \\
Sonar         &   208 & 60 &  2 & KEEL   &   89 & 0.524 \\
Ionosphere    &   351 & 33 &  2 & KEEL   &   99 & 0.686 \\
WDBC          &   569 & 30 &  2 & KEEL   &   93 & 0.819 \\
Segment       &  2310 & 19 &  7 & KEEL   &  396 & 0.809 \\
Spambase      &  4597 & 57 &  2 & KEEL   & 1227 & 0.703 \\
Waveform      &  5000 & 40 &  3 & OpenML & 2600 & 0.422 \\
Banana        &  5300 &  2 &  2 & KEEL   & 1177 & 0.753 \\
Phoneme       &  5404 &  5 &  2 & KEEL   & 1423 & 0.707 \\
Texture       &  5500 & 40 & 11 & KEEL   &  866 & 0.825 \\
Satimage      &  6435 & 36 &  6 & KEEL   & 1549 & 0.733 \\
Ring          &  7400 & 20 &  2 & KEEL   & 2915 & 0.562 \\
Twonorm       &  7400 & 20 &  2 & KEEL   & 1006 & 0.849 \\
Penbased      & 10992 & 16 & 10 & KEEL   & 1052 & 0.894 \\
EEG Eye State & 14980 & 14 &  2 & UCI    & 6369 & 0.528 \\
Magic         & 19020 & 10 &  2 & KEEL   & 7214 & 0.579 \\
Letter        & 20000 & 16 & 26 & KEEL   & 6816 & 0.621 \\
\midrule
Mean          &       &    &    &        & 2055 & 0.692 \\
\bottomrule
\end{tabular}

\end{table*}

\subsection{Method Evaluations}
Before any fitting, SPINE reserves a stratified quarter of each training fold. Every risk-based decision is then evaluated on that held-out portion. That is, edge screening in Section~\ref{sec:phase2a}, the choice of growth site in Section~\ref{sec:phase3} (with the exception of the Phase 3 fallback which scores by candidate sites), and the pruning order in Section~\ref{sec:phase4} are all scored there. No decision at any stage sees the test partition. SPINE therefore builds its model from 75\% of each training fold, while every competitor uses all of it.

Each dataset is scored by its mean accuracy over the ten test folds per each dataset. Full represents the accuracy of a 1-NN classifier trained on the entire training fold. It serves as an accuracy reference and does not enter the statistical comparisons. We apply the Friedman test across SPINE and the competing methods. Upon rejection, we compare SPINE against each competitor using the Wilcoxon signed-rank test, and we also report the Conover post-hoc procedure \citep{conover_multiple-comparisons_1979}. Both are adjusted across the seven comparisons by Holm's procedure \citep{holm_simple_1979} at the $\alpha = 0.05$ level. Datasets on which SPINE and a competitor tie exactly are dropped from the signed-rank statistic, and we report the win, loss, and tie counts alongside each $p$-value. 

\subsection{Results}
Table~\ref{tab:accuracy} reports mean accuracy over the ten outer folds. SPINE attains the highest mean across the seventeen datasets at 0.9110, ahead of LVQ3 at 0.9015 and RSP3 at 0.8915. It is the most accurate compared method on six datasets, and it places among the top three on thirteen. Full exceeds every compared method on Segment, Phoneme, Texture, and Penbased. The table also shows a difference in consistency. The worst standing SPINE reaches is sixth of eight on Penbased, and its largest shortfall from the best compared method on a dataset is 0.0332 on Waveform. Every other method except LVQ3 finishes last on at least one dataset. 

The wins are not spread evenly. All six datasets on which SPINE is best have two classes. The two largest margins are on Banana at $+0.0225$ and Ring at $+0.0176$, both of which have curved class supports. Across the ten binary datasets the mean rank of SPINE is 1.70, and across the seven multiclass datasets it is 3.50. SPINE remains among the top three on five of the seven multiclass datasets, so the pattern is a loss of margin rather than a failure. The matched budget is apportioned over classes, so a fixed total is divided more finely as the class count grows, which may leave each skeleton with too few vertices to describe its class. Note that Letter has twenty-six classes and SPINE ranks second there, so the effect does not follow the class count monotonically. SPINE exceeds Full on twelve datasets, trails it on four, and ties on Wine. The mean difference is $+0.0155$ in favor of SPINE. The four datasets on which it trails are the ones named above, and three of them are multiclass. A reduced set that matches the full training fold is not unusual under 1-NN, since discarding points may remove mislabeled and boundary noise.

Across the campaign the frozen Betti numbers from Phase 2a survived to emission on all 860 class-folds. The Phase 3 cap did not bind on any class of any fold. The closest approach was on Waveform, where the overshoot target reached 94.6\% of the class's points in the fitting portion. No removal was forced on any fold. The soundness check, that no class ends with fewer components or more cycles than its nerve, was applied at every emission and never failed. Edge screening is thus the only stage that moved the numbers. It removed 869 edges across the campaign. Those removals raised the mean $\beta_0$ per class from 1.05 at the nerve to 2.05 at the
freeze, and lowered the mean $\beta_1$ from 0.050 to 0.031. Note that $\beta_1$ is small because the nerves are nearly always forests. Only 35 of the 860 class-folds carry any cycle at the nerve, 43 in all, and 26 do at emission, 27 in all. Those 26 fall on Segment, Phoneme, Penbased, and Spambase alone. What is preserved on this roster is therefore mostly the component count. 

\begin{table*}[t]
\centering
\small
\caption{Mean 1-NN accuracy over the ten outer folds. Full retains the training fold and is an accuracy reference rather than a competing method, so it is excluded from the ranks and from the statistical comparison. Every method except Full emits the matched budget $M$ of Table~\ref{tab:datasets}. Bold marks the highest accuracy among the compared methods on each dataset, with Full excluded. Average rank is taken over the seventeen datasets, lower being better.}
\label{tab:accuracy}
\begin{tabular}{lccccccccc}
\toprule
Dataset & Full & SPINE & RSP3 & KMeans & Random & LVQ3 & SPOT & GLVQ & GNG \\
\midrule
Wine          & 0.9552 & 0.9552 & 0.9386 & 0.9552 & 0.9493 & 0.9497 & 0.9497 & \textbf{0.9716} & 0.9556 \\
Sonar         & 0.8655 & 0.8931 & 0.8745 & 0.8790 & 0.8317 & 0.8843 & 0.8557 & 0.6931 & \textbf{0.8979} \\
Ionosphere    & 0.8660 & 0.8860 & 0.8833 & 0.8062 & 0.8319 & 0.8719 & \textbf{0.8888} & 0.8661 & 0.8635 \\
WDBC          & 0.9472 & 0.9525 & 0.9384 & 0.9578 & 0.9366 & \textbf{0.9666} & 0.9578 & 0.9314 & \textbf{0.9666} \\
Segment       & 0.9623 & 0.9455 & \textbf{0.9610} & 0.9450 & 0.9143 & 0.9515 & 0.9385 & 0.8866 & 0.9450 \\
Spambase      & 0.9139 & \textbf{0.9143} & 0.8884 & 0.9041 & 0.8762 & 0.9126 & 0.8804 & 0.8997 & 0.5832 \\
Waveform      & 0.7148 & 0.7810 & 0.7584 & 0.7382 & 0.7050 & 0.7608 & 0.7120 & \textbf{0.8142} & 0.7506 \\
Banana        & 0.8758 & \textbf{0.9032} & 0.8455 & 0.8491 & 0.8711 & 0.8808 & 0.8685 & 0.8687 & 0.8570 \\
Phoneme       & 0.9060 & \textbf{0.8910} & 0.8705 & 0.8793 & 0.8494 & 0.8834 & 0.8608 & 0.7994 & 0.8877 \\
Texture       & 0.9891 & 0.9785 & 0.9829 & 0.9824 & 0.9622 & 0.9811 & 0.9769 & 0.9180 & \textbf{0.9838} \\
Satimage      & 0.9035 & 0.9038 & 0.9029 & 0.8988 & 0.8855 & \textbf{0.9110} & 0.8875 & 0.8483 & 0.9050 \\
Ring          & 0.7538 & \textbf{0.8574} & 0.8141 & 0.7335 & 0.7234 & 0.7878 & 0.8280 & 0.7542 & 0.8399 \\
Twonorm       & 0.9461 & 0.9758 & 0.9251 & 0.9588 & 0.9378 & 0.9634 & 0.9507 & \textbf{0.9777} & 0.9630 \\
Penbased      & 0.9945 & 0.9841 & 0.9909 & 0.9912 & 0.9784 & \textbf{0.9922} & 0.9901 & 0.9404 & 0.9918 \\
EEG Eye State & 0.8565 & \textbf{0.8658} & 0.8388 & 0.8479 & 0.8284 & 0.8509 & 0.8355 & 0.7590 & 0.8518 \\
Magic         & 0.8183 & \textbf{0.8446} & 0.7909 & 0.8101 & 0.8055 & 0.8292 & 0.8115 & 0.8048 & 0.8172 \\
Letter        & 0.9547 & 0.9555 & 0.9507 & 0.9538 & 0.9197 & 0.9483 & 0.9440 & 0.7894 & \textbf{0.9559} \\
\midrule
Mean          & 0.8955 & 0.9110 & 0.8915 & 0.8877 & 0.8710 & 0.9015 & 0.8904 & 0.8543 & 0.8833 \\
Average Rank  &   ---  & 2.44 & 4.94 & 4.62 & 6.82 & 2.94 & 5.18 & 5.94 & 3.12 \\
\bottomrule
\end{tabular}
\end{table*}

The Friedman test rejects the hypothesis that the eight methods are equivalent across the seventeen datasets ($\chi^{2} = 47.54$ on 7 degrees of freedom, $p = 4.4 \times 10^{-8}$). SPINE attains the best average rank at 2.44, ahead of LVQ3 at 2.94 and GNG at 3.12. Table~\ref{tab:stats} reports the seven comparisons against SPINE. Five methods (Random, SPOT, GLVQ, RSP3, and KMeans) reject at $\alpha = 0.05$ under both procedures. Note that the KMeans comparison uses sixteen datasets rather than seventeen, since SPINE and KMeans score identically on Wine. The comparisons between LVQ3 and GNG do not reach significance at the $\alpha = 0.05$ level. LVQ3 and GNG both place prototypes under an objective of their own. The win counts point the same way, at 12 to 5 against LVQ3 and 10 to 7 against GNG. 

\begin{table}[t]
\centering
\small
\caption{Wilcoxon signed-rank test of SPINE against each competitor on mean 1-NN accuracy over the seventeen datasets. Counts of wins, losses and ties (W/L/T) are for SPINE. Comparisons is the number of datasets entering the signed-rank statistic, which excludes exact ties. Both columns of adjusted $p$-values are corrected by Holm across these seven comparisons. The median difference of SPINE minus the competitor is reported.}
\label{tab:stats}
\begin{tabular}{lccccc}
\toprule
& & & \multicolumn{2}{c}{Holm-adjusted $p$} & \\
\cmidrule(lr){4-5}
Comparison & W/L/T & Comparisons & Wilcoxon & Conover & Median Difference \\
\midrule
SPINE vs Random & 17/0/0 & 17 & 0.0001 & $<0.0001$ & 0.0373 \\
SPINE vs GLVQ   & 14/3/0 & 17 & 0.0039 & $<0.0001$ & 0.0437 \\
SPINE vs SPOT   & 14/3/0 & 17 & 0.0039 & 0.0004    & 0.0251 \\
SPINE vs RSP3   & 14/3/0 & 17 & 0.0084 & 0.0012    & 0.0186 \\
SPINE vs KMeans & 13/3/1 & 16 & 0.0101 & 0.0046    & 0.0116 \\
SPINE vs LVQ3   & 12/5/0 & 17 & 0.0611 & 0.6295    & 0.0076 \\
SPINE vs GNG    & 10/7/0 & 17 & 0.0797 & 0.6295    & 0.0033 \\
\bottomrule
\end{tabular}
\end{table}

\section{Discussions} \label{sec:discussion}

The results of Section~\ref{sec:results} score SPINE under its own decision rule, which is a larger hypothesis class than the competitors. The same fitted model can also be scored as a vertex-only prototype set, by discarding the segments at prediction time and classifying by the nearest vertex. In this section we examine the effects of the graph segments on classification as well as evaluate the construction costs relative to the methods it competes with. 

\subsection{Effect of the Graph Segments and Budget} \label{sec:readout}

The emitted model can be evaluated under the decision rule of Equation~\eqref{eq:spine-rule} or using just the vertices. We write SPINE+Graph for the first and SPINE+1NN for the second. The two differ in the decision rule and in nothing else, since they come from the same fit, one set of vertices, and one seed. Their paired differences therefore isolate the changes of the hypothesis class. Under the same evaluation framework, using SPINE+1NN instead of SPINE+Graph, the Friedman test rejects ($\chi^{2} = 44.94$ on 7 degrees of freedom, $p = 1.4 \times 10^{-7}$). SPINE+1NN holds the best average rank at 2.79, narrowly ahead of LVQ3 at 2.82. The set of competitors it separates from is unchanged. Wilcoxon sign-rank test reject the same five methods, namely Random, SPOT, GLVQ, RSP3, and KMeans. 

At the matched budget the average difference is small, at 0.9110 for SPINE+Graph against 0.9080 for SPINE+1NN with a median difference of 0.0007. SPINE+Graph wins on ten datasets, loses on four, and ties on three. Wilcoxon signed-rank test over the retained differences gives $p = 0.091$, so the advantage do not reach significance at the $0.05$ level.

The arguments of Section~\ref{sec:intro} predicts using edge segments matter more when vertices are scarcer, since an edge covers the continuum at the cost of two stored points. RSP3's count retains about $31\%$ of the training fold on average. We therefore sweep the budget across a grid that reaches well below it on average. Each of the budget methods are rebuilt at six fractions of the training fold size from $0.5\%$ to $25\%$, and scored on 1-NN accuracy. Table~\ref{tab:readout} reports the paired difference between the two SPINE decision rules at each fraction. The direction is consistent across the grid, and using the full complex wins on at least ten datasets at each budget point. The median difference increases monotonically from $0.0009$ at $25\%$ to $0.0056$ at $0.5\%$. That is the pattern the covering argument predicts, and it is the strongest evidence we have for the segments.

\begin{table}[t]
\centering
\small
\caption{The paired difference in average 1-NN accuracy between SPINE+Graph minus SPINE+1NN across the budget grid over the seventeen datasets. The budget at each grid point is given. Positive values favor the skeleton rule. Wins, losses, and ties are reported from SPINE+Graph.}
\label{tab:readout}
\begin{tabular}{lrrrrrr}
\toprule
Budget fraction & 0.5\% & 1\% & 2\% & 5\% & 10\% & 25\% \\
\midrule
Mean difference   & 0.0208 & 0.0108 & 0.0061 & 0.0075 & 0.0049 & 0.0023 \\
Median difference & 0.0056 & 0.0042 & 0.0041 & 0.0036 & 0.0020 & 0.0009 \\
Win / loss / tie  & 12/2/3 & 14/1/2 & 11/5/1 & 15/2/0 & 15/2/0 & 10/5/2 \\
\bottomrule
\end{tabular}
\end{table}

\begin{figure}
  \centering
  \includegraphics[width=.75\linewidth]{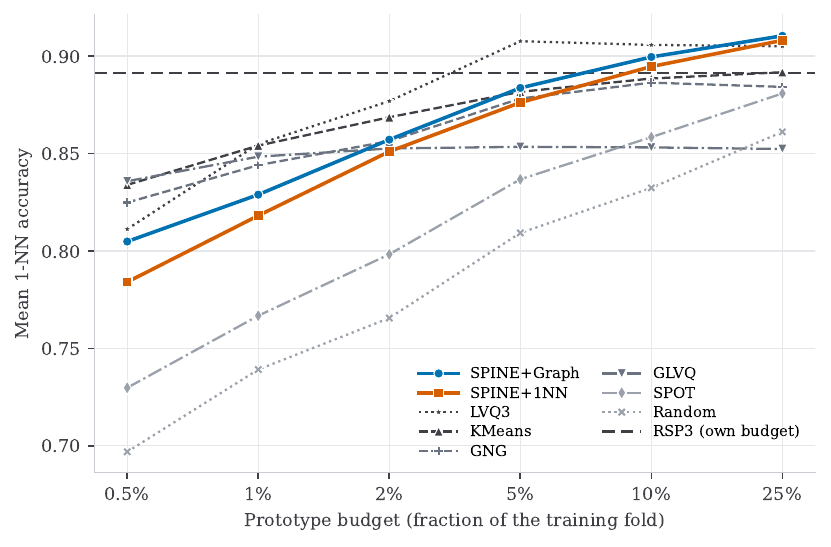}
  \caption{Average mean 1-NN accuracy against the prototype budget. The budget at each grid point is the same fraction of the training fold for every swept method. RSP3 is built once per fold and drawn as a horizontal reference at its own operating point. That point averages $31\%$ of the training fold, beyond the right edge of the grid.}
  \label{fig:curve}
\end{figure}

The sweep also locates the regime in which SPINE competes. Figure~\ref{fig:curve} shows both SPINE decision rules against the other methods across the grid. At $25\%$ of the training fold SPINE leads at $0.9103$, ahead of LVQ3 at $0.9050$ and KMeans at $0.8917$. At $0.5\%$ it places fifth among the swept methods at $0.8048$, behind GLVQ at $0.8357$ and KMeans at $0.8339$. SPINE passes the RSP3 accuracy between $5\%$ and $10\%$ at roughly a third of the budget RSP3 chooses for itself, and passes LVQ3 between $10\%$ and $25\%$. That is, the graph segments buy more at small budgets while the method as a whole competes best at more moderate budgets. Phase 0 fixes the nerve before the budget is known, so a budget below the size of the nerve forces removals the later phases cannot avoid. 

\subsection{Construction Cost} \label{sec:timing}

Construction cost is measured in a separate study, five folds per dataset. All timings are taken on an Apple M2 chip with 8 core CPU and 32 GB of memory. BLAS is restricted to a single thread. The roster separates into two groups. SPINE sits within the discriminative and graph-building methods, instead of the selection and unconstrained quantize methods. Scaling is the organizing fact. We fit a power law with exponent $b$ to mean construction time against training-fold size on log-log axes.  

\begin{figure*}[t]
  \centering
  \includegraphics[width=\linewidth]{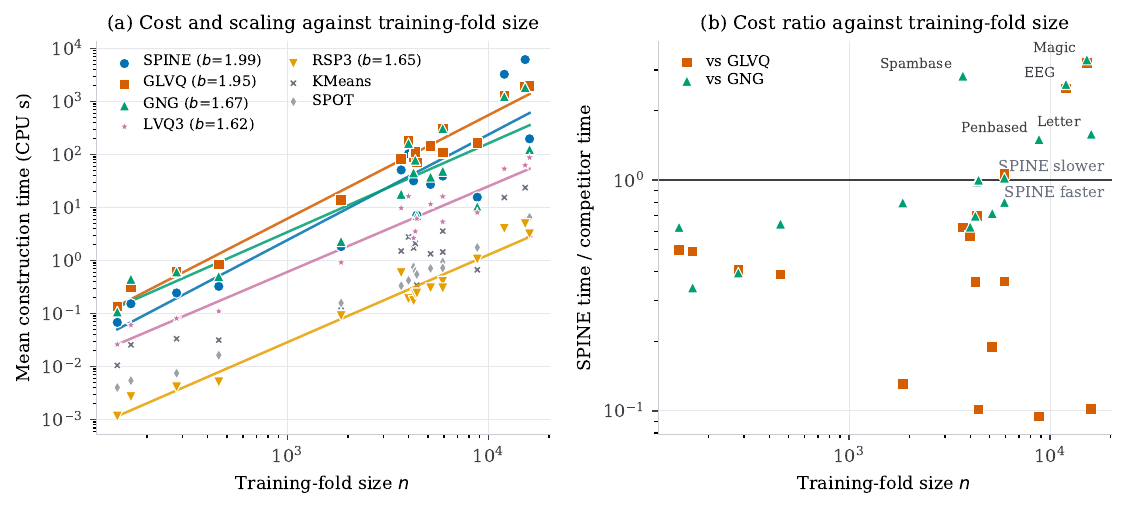}
  \caption{Construction cost, from the fixed-configuration five-fold timing study. (a) Mean construction time against training-fold size on log-log axes, with a fitted power law for each method Table~\ref{tab:timing} reports. The legend gives the fitted exponent $b$. Cost separates into
  a two groups comprising the selection methods and the unconstrained quantizer as one group, and discriminative and graph-building methods, in which SPINE sits, as the other. (b) SPINE's cost relative to GLVQ and to GNG against training-fold size, with the line of equality. Points below the line are datasets on which SPINE is the faster of the two, and the datasets on which it is slower than GNG by more than a factor of 1.4 are named. The ratio against GLVQ has no trend in fold size, while the ratio against GNG rises with it. Color marks the methods Table~\ref{tab:timing} reports, with
  KMeans and SPOT in gray.}
  \label{fig:timing}
\end{figure*}

The exponents are 1.99 for SPINE, 1.95 for GLVQ, 1.67 for GNG, 1.65 for RSP3 and 1.62 for LVQ3. SPINE and GLVQ therefore grow at similar rates. That matters, because GLVQ descends the same margin over isolated prototypes and is the closest comparison to SPINE. SPINE is faster than GLVQ on fourteen of the seventeen datasets, with a median ratio of 0.41 and no trend in fold size. It is slower than GLVQ marginally on Ring, at 1.06 times, and materially on EEG Eye State and Magic, at 2.5 and 3.2 times. Those two datasets have two classes each and the largest per-class vertex counts across other datasets, near 3000. Letter carries a similar total budget across 26 classes, near 240 vertices per class, and there SPINE runs at a tenth of GLVQ. That is, the cost is driven by the number of vertices in a single class rather than by the budget itself. Against GNG the comparison is less favorable than the median suggests. The median ratio is 0.80, however the ratio rises with fold size and reaches 2.6 on EEG Eye State and 3.3 on Magic. The Mapper construction and the 1-complex are therefore not free relative to a graph-building quantizer, and the gap widens with the training fold. Table~\ref{tab:timing} reports the means of SPINE, GLVQ, GNG, LVQ3 and RSP3. Figure~\ref{fig:timing} shows the fitted scaling in panel (a) and both cost ratios against fold size in panel (b).

\begin{table}[t]
\centering
\small
\caption{Mean construction time in seconds, from the fixed-configuration five-fold timing study on a single BLAS thread. All methods share the budget $M$ within this study, however the five-fold split leaves a smaller training fold than the ten-fold protocol, so $M$ here is not the $M$ of Table~\ref{tab:datasets}. Datasets are ordered by training-fold size. KMeans and SPOT are omitted; both sit with RSP3 in the fast group, and neither exceeds RSP3 by more than a factor of fifteen.}
\label{tab:timing}
\begin{tabular}{lrrrrrrr}
\toprule
Dataset & $n_{\text{train}}$ & $M$ & SPINE & GLVQ & GNG & LVQ3 & RSP3 \\
\midrule
Wine          &   142 &   34 &    0.07 &    0.14 &    0.11 &  0.03 & 0.00 \\
Sonar         &   166 &   79 &    0.15 &    0.31 &    0.45 &  0.06 & 0.00 \\
Ionosphere    &   281 &   94 &    0.25 &    0.60 &    0.62 &  0.08 & 0.00 \\
WDBC          &   455 &   79 &    0.33 &    0.84 &    0.50 &  0.11 & 0.01 \\
Segment       &  1848 &  374 &    1.84 &   14.04 &    2.30 &  0.92 & 0.09 \\
Spambase      &  3678 & 1109 &   51.05 &   82.07 &   18.02 &  9.79 & 0.59 \\
Waveform      &  4000 & 2280 &  103.13 &  179.84 &  164.47 & 16.32 & 0.19 \\
Banana        &  4240 & 1049 &   31.80 &   88.14 &   45.47 &  2.67 & 0.17 \\
Phoneme       &  4323 & 1315 &   77.92 &  111.50 &   78.98 &  3.54 & 0.23 \\
Texture       &  4400 &  807 &    7.17 &   71.04 &    7.16 &  6.16 & 0.24 \\
Satimage      &  5148 & 1412 &   27.23 &  144.13 &   37.94 & 11.59 & 0.30 \\
Ring          &  5920 & 2597 &  320.32 &  301.66 &  312.81 & 16.10 & 0.40 \\
Twonorm       &  5920 &  921 &   39.26 &  108.15 &   48.89 &  5.38 & 0.30 \\
Penbased      &  8794 &  951 &   15.64 &  165.59 &   10.42 &  8.04 & 1.05 \\
EEG Eye State & 11984 & 5539 & 3258.51 & 1304.76 & 1250.40 & 53.63 & 3.99 \\
Magic         & 15216 & 6431 & 6159.48 & 1908.05 & 1848.77 & 62.53 & 4.91 \\
Letter        & 16000 & 6203 &  197.41 & 1943.27 &  124.49 & 87.35 & 3.16 \\
\bottomrule
\end{tabular}
\end{table}

\section{Conclusion} \label{sec:conclusion}

In this paper we introduce a prototype generation method using skeletal prototypes on iterative nerve expansions, called SPINE where each class is an embedded 1-complex representing the skeleton of the class points. The vertices are the prototypes, and the segments joining them enter the decision rule rather than only the fitting. The initial edge set comes from class-conditional Mapper constructions. The data therefore decides which localized clusters are joined, at a resolution set by the size of the class. Every later phase fits that structure under a classification objective. To our knowledge, this is the first use of Mapper for supervised prototype reduction. 

Across seventeen datasets at a matched budget, SPINE attains the highest mean accuracy at $0.9110$. SPINE also exceeds the accuracy trained on the full training data on twelve datasets. It also holds the best average rank of the eight methods compared, at $2.44$. The Friedman test rejects equivalence, and under Wilcoxon signed-rank test with Holm correction SPINE is significantly better than five of the seven competitors. It does not reach significance against LVQ3 and GNG, and the Conover procedure agrees on every comparison. These results are obtained while SPINE builds its model from a portion of the training data, using the other portion as validation holdout while other methods use the entire training fold. The wins concentrate on the binary datasets, where the average rank is $1.70$ against $3.50$ on the multiclass ones. 

Scoring the same fitted model on the vertices only prototype set isolates the effects of the graph segments. At the matched budget that difference is small and does not reach significance. We sweep SPINE under both decision rules across different training set budgets. Across the grid the median gain of using the entire graph rises monotonically as the budget falls, from $0.0009$ at a budget of $25\%$ to $0.0056$ at $0.5\%$. That is the pattern the covering argument predicts, since an edge covers a continuum at the cost of two stored points. The method as a whole behaves in the opposite direction. SPINE leads the roster at $25\%$ of the training fold and places fifth at $0.5\%$. We therefore recommend using SPINE over competitors at moderate budgets, noting that SPINE passes RSP3 at a budget of $10\%$ and LVQ3 at $25\%$. 

The Betti numbers of the original Mapper complex are stored for comparison in Phase 2a, and can be checked at the emitted model rather than assumed. That separates SPINE from a method that builds a graph while fitting and then discards it. What the check protects is modest on this roster, however. Edge screening is the only stage that moved the Betti numbers, and the nerves are nearly always forests. The quantity preserved is therefore mostly the component count, since cycles survive on 26 out of 860 class-folds and fall on four datasets alone. The guarantee is also budget-dependent, as Phase 0 fixes the nerve before the budget is consulted. 

Construction cost places SPINE with the discriminative and graph-building methods such as GLVQ and GNG. SPINE and GLVQ scale at the same rate in the training-fold size, and SPINE is the faster of the two on fourteen out of the seventeen datasets. Against GNG the cost ratio rises with fold size, so the Mapper construction and the complex are not free relative to a graph-building quantizer.

\subsection{Limitations and Future Work}
We note that the lens, the clusterer, the admission criterion, the step rule, and the boundary restriction were held at declared settings throughout. Those settings were fixed in advanced as heuristics rather than selected by cross-validation, which keeps the comparison from selecting on the evaluation suite. The cost is that the results report one configuration, and the sensitivity of the method to that configuration is untested. 

While many of these heuristics were chosen intuitively, they are ultimately not theoretically derived. The interval count of Equation~\eqref{eq:n-intervals}, the gain $g = 0.25$, the two HDBSCAN constants $\kappa$ and $\lambda$, the squashing scale $\theta$ of Equation~\eqref{eq:loss} and the five schedule constants of Equation~\eqref{eq:phase1-schedule} were each declared in advance with $\sigma_0$ set from the nerve sizes of Phase 0 produces, and we make no optimality claim for any of them. Their sensitivities as such are also not reported in this work. A theoretical account of all parameters, with the resolution rule and the gain since together they decide what adjacency means in the initial complex, is the largest gap we leave open. 

We note that SPINE leads strongly on binary classification problems, but worsens under multiclass classification. The construction builds each class independently, so a multiclass problem is treated as a collection of one-class problems. The matched budget is then apportioned over classes, and a fixed total is divided more finely as the class count grows. An extension in which the classes inform one another, either in the cover or in the apportionment, is the natural extension.

The growth phase of Phase 3 falls back on representation error whenever no held-out error can be attributed to a site, which is a discrete rule standing in for a continuous one. A continuous site score would remove the fallback, and potentially could also change the cost of the phase, since the held-out evaluation is what the current rule spends its time on. 

The cover is fixed before the budget is consulted, so a budget below the size of the nerve leaves the later phases no homotopy-preserving removal. That is part of why the method places fifth at the smallest budget on the grid. A budget-aware Phase 0, in which the resolution rule takes the budget as an argument, would remove that failure mode.

\bibliographystyle{cas-model2-names}

\bibliography{references}


\end{document}